%% file: main_arxiv.tex
\documentclass{article}

\usepackage[preprint]{neurips_2023}

\usepackage[utf8]{inputenc} 
\usepackage[T1]{fontenc}    
\usepackage{hyperref}       
\usepackage{url}            
\usepackage{booktabs}       
\usepackage{amsfonts}       
\usepackage{nicefrac}       
\usepackage{microtype}      
\usepackage{xcolor}         

\usepackage{graphicx}
\usepackage{colortbl}
\usepackage{tabulary}
\usepackage{makecell}
\usepackage{threeparttable}
\usepackage{wrapfig}

\usepackage{amsmath}
\usepackage{multirow}

\usepackage{pifont}
\newcommand{\cmark}{\ding{51}}%

\newcommand{\methodname}{\textsc{SELMA}}

\usepackage[capitalize]{cleveref}
\crefname{section}{Sec.}{Secs.}
\Crefname{section}{Section}{Sections}
\Crefname{table}{Table}{Tables}

\crefname{table}{Table}{Tables}

\usepackage{eccvabbrv}

\setcitestyle{square,numbers}
\hypersetup{colorlinks,linkcolor={magenta},citecolor={blue},urlcolor={magenta}}

\title{SELMA:
Learning and Merging
Skill-Specific
Text-to-Image
Experts
with Auto-Generated Data}

\newcommand*\samethanks[1][\value{footnote}]{\footnotemark[#1]}

\author{%
Jialu Li\thanks{equal contribution}
\;\;\;\;
Jaemin Cho\samethanks{}
\;\;\;\;
Yi-Lin Sung
\;\;\;\;
Jaehong Yoon
\;\;\;\;
Mohit Bansal \\
  UNC Chapel Hill \\
  \texttt{\{jialuli, jmincho, ylsung, jhyoon, mbansal\}@cs.unc.edu} \\ \\
  \url{https://selma-t2i.github.io}
}
\begin{document}

\maketitle

\begin{abstract}
Recent text-to-image (T2I) generation models have demonstrated impressive capabilities in creating images from text descriptions.
However, these T2I generation models often fall short of generating images that precisely match the details of the text inputs, such as incorrect spatial relationship or missing objects. In this paper, we introduce \textbf{\methodname{}}: \textbf{S}kill-Specific \textbf{E}xpert \textbf{L}earning and \textbf{M}erging with \textbf{A}uto-Generated Data, a
novel paradigm to improve the faithfulness
of T2I models by fine-tuning models on
automatically generated, multi-skill
image-text datasets, 
with skill-specific expert learning and merging.
First, \methodname{} leverages an LLM's in-context learning capability to generate multiple datasets of text prompts that can teach different skills, and then generates the images with a T2I model based on the prompts.
Next, \methodname{} adapts the T2I model to the new skills
by learning multiple single-skill LoRA (low-rank adaptation) experts followed by expert merging.
Our independent expert fine-tuning specializes multiple models for different skills,
and expert merging
helps build a joint multi-skill T2I model that can generate faithful images given diverse text prompts,
while mitigating the knowledge conflict
from different datasets. 
We empirically demonstrate that \methodname{} significantly improves the semantic alignment and text faithfulness of state-of-the-art T2I diffusion models on multiple benchmarks ($+$2.1\% on TIFA and $+$6.9\% on DSG),
human preference metrics (PickScore, ImageReward, and HPS),
as well as human evaluation.
Moreover,
fine-tuning with image-text pairs auto-collected via \methodname{}
shows comparable performance to fine-tuning with ground truth data.
Lastly, we show that fine-tuning with images from a weaker T2I model can help improve the generation quality of a stronger T2I model, suggesting promising weak-to-strong generalization in T2I models.
\end{abstract}

\section{Introduction}
\label{sec:intro}
Text-to-Image (T2I) generation models have shown impressive development in recent years~\cite{saharia2022photorealistic,rombach2022high,podell2023sdxl,Kang2023ScalingUG,ramesh2021zero,yu2022scaling,chang2023muse}.
Although these approaches can generate high-quality, diverse images with the unseen composition of objects (\eg{}, Eiffel Tower under water)
based on textual inputs, they still struggle to capture and reflect all semantics in the given textual prompts. In particular, the state-of-the-art T2I models are known to often fail to generate multiple subjects~\cite{yu2022scaling, feng2022training, liu2022compositional}, spatial relationships between entities~\cite{phung2023grounded}, and text rendering~\cite{liu2022character, tuo2023anytext} (\eg{}, an art board with ``SELMA'' written on it) in the prompts.

\begin{figure}[t!]
    \centering
    \includegraphics[width=\textwidth]{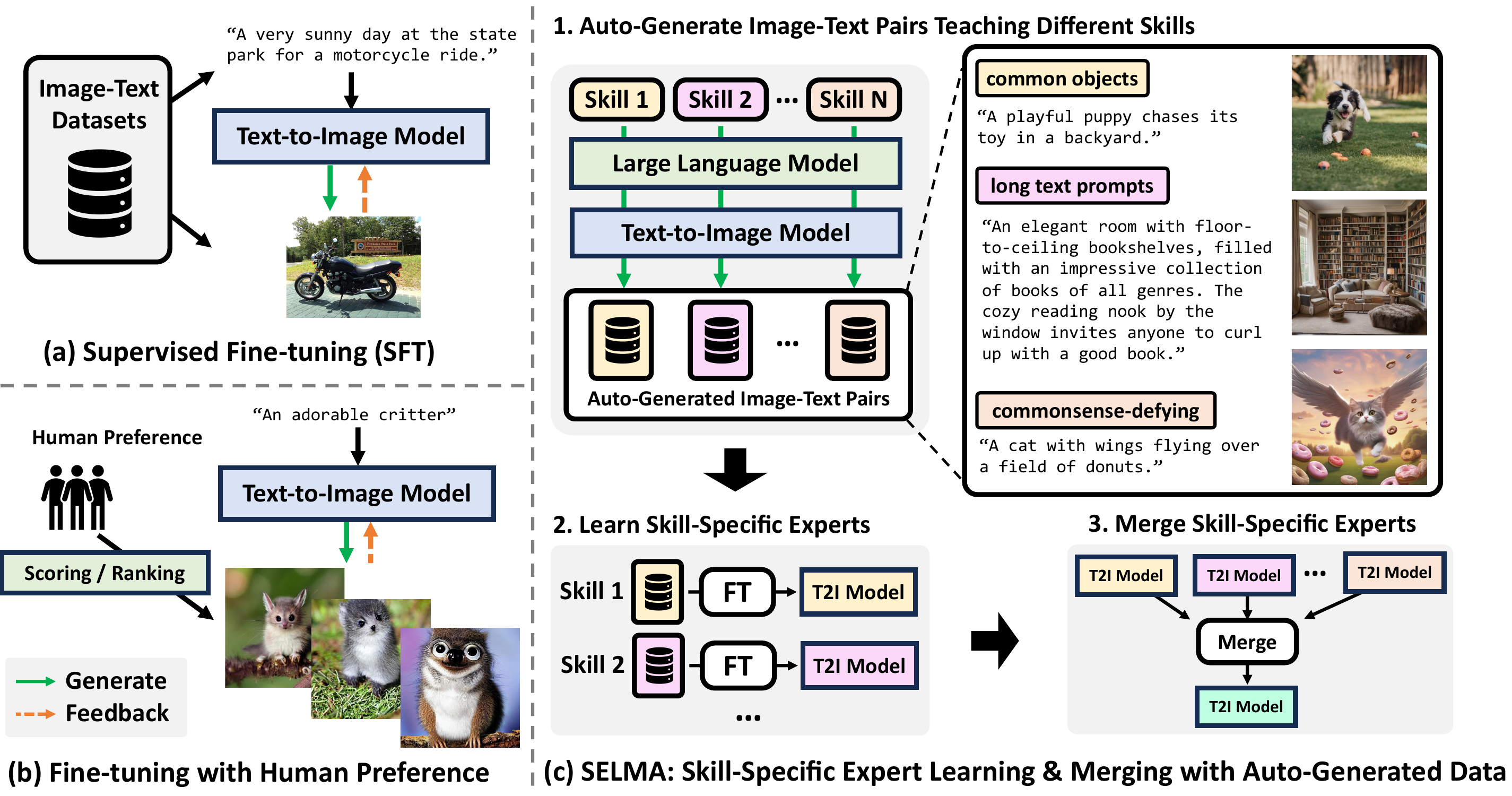}
    \caption{
    Comparison of different fine-tuning paradigms for text-to-image (T2I) generation models.
    \textbf{(a) Supervised Fine-tuning (SFT)}: a T2I model is trained with image-text pairs from existing datasets.
    \textbf{(b) Fine-tuning with Human Preference (\eg{}, RL/DPO)}: 
    humans annotate their preferences on images by ranking/scoring in terms of text alignments,
    and a T2I model is trained to maximize the human preference scores.
    \textbf{(c) \methodname{}}:
    instead of collecting image-text pairs or human preference annotations,
    we automatically collect image-text pairs for desired skills with LLM and T2I model,
    and create a multi-skill T2I model by
    learning and merging skill-specific expert models. 
    }
   \label{fig:teaser}
\end{figure}

Many recent works have been proposed to tackle these challenges in text-to-image generation, aiming to enhance the faithfulness of T2I models to textual inputs.
One line of research focuses on supervised fine-tuning
on high-quality image-text datasets with human annotations~\cite{dai2023emu} or image-text pairs with re-captioned text prompts~\cite{segalis2023picture, betker2023improving}, as shown in \cref{fig:teaser} (a).
Another line of research is based on aligning T2I models with human preference annotations~\cite{xu2023imagereward, prabhudesai2023aligning, fan2023dpok, lee2023aligning, wallace2023diffusion}, as shown in \cref{fig:teaser} (b).
Other works focus on introducing additional layouts or object grounding boxes to guide the generation process~\cite{li2023gligen, xie2023boxdiff, yang2023reco, feng2022training, cho2023visual, zhang2023adding}.
Despite achieving significant improvements in aligning generated images with input textual prompts,
the success of these approaches relies on the quality of the layouts created from the textual prompts,
the collection of high-quality annotations with human efforts, or the existence of large-scale ground truth data,
which involves expensive human annotation.

Motivated by LLMs' impressive text generation capability (given open-ended task instructions and in-context examples), and recent T2I models' capability in generating highly realistic photos (based on text prompts), we investigate an interesting question to further improve the faithfulness of state-of-the-art T2I models: ``\emph{Can we automatically generate multi-skill image-text datasets with LLMs and T2I models, to effectively and efficiently teach different image generation skills to T2I models?}''
In this paper, we propose \textbf{\methodname{}}: \textbf{S}kill-Specific \textbf{E}xpert \textbf{L}earning and \textbf{M}erging with \textbf{A}uto-Generated Data,
a novel paradigm for eliciting the pre-trained knowledge in T2I models for improved faithfulness based on skill-specific learning and merging of experts.
\methodname{} consists of four stages: 
(1) collecting skill-specific prompts with in-context learning of LLMs,
(2) self-generating image-text samples for diverse skills without the need of human annotation nor feedback from reward models,
(3) fine-tuning the expert T2I models on these datasets separately,
and
(4) obtaining the final model by merging experts of each dataset for efficient adaptation to different skills and mitigation of knowledge conflict in joint training.
We illustrate the \methodname{} pipeline in \cref{fig:teaser} (c).

In the first and second stages, we use the LLM and the T2I model to generate skill-specific image-text data (which will be used for fine-tuning T2I models in later stages). We aim to teach diverse generation skills to the same T2I model (\ie{}, self-learning), so that they can handle different types of prompts.
For example,
some prompts require
object counting capability of T2I models (\eg{}, ``\textit{Two leaves and two wallets}''), while other prompts specify a list of desired attributes in the generated image (\eg{}, ``\textit{high priestess tarot card, black background, moon theme, witchery, great detail, sketch art with intricate background, dynamic pose, closeup}'').
To generate image-text pairs for different skills,
we first query GPT-3.5~\cite{gpt35} for prompt generation by using only three skill-specific prompts as in-context examples, and filter the generated prompts with ROUGE-L score to maximize prompt diversity to collect 1K prompts in total (\cref{sec:sub:prompt_collect}).
Then we use Stable Diffusion models~\cite{rombach2022high,podell2023sdxl} themselves to generate corresponding images from the prompts (\cref{sec:sub:image_collect}).
We find that our skill-specific training
can help
mitigate
knowledge conflict
when jointly learning multiple skills (see \Cref{table:lora_merging}).

In the third and fourth stages, we fine-tune a T2I model with the collected image-text pairs to teach different skills. However, updating the entire model weights
can be inefficient; knowledge conflicts within mixed datasets may also lead to suboptimal performance~\cite{Liu2019LossBalancedTW}.
Thus, in the third stage, we fine-tune T2I models on these self-generated image-text pairs with parameter-efficient LoRA (low-rank adaptation) modules~\cite{hu2021lora} 
to create skill-specific expert T2I models
(\cref{sec:sub:finetuning_lora}).
In the fourth stage, to build a joint multi-skill T2I model that can have faithful generations across different skills,
we merge the skill-specific experts based on LoRA merging~\cite{Shah2023ZipLoRAAS,zhong2024multi}
(\cref{sec:sub:lora_merging}).
We find that our inference-time merging of skill-specific LoRA experts
is effective in mitigating skill conflicts than training a mixture of LoRA expert model~\cite{wu2023mole}
while also being more efficient (see \Cref{table:training_ablation}).

We validate the usefulness of \methodname{} with
public
state-of-the-art
T2I models -- a family of Stable Diffusion -- v1.4~\cite{rombach2022high}, v2~\cite{rombach2022high}, and XL~\cite{podell2023sdxl}
on two text faithfulness evaluation benchmarks (DSG~\cite{cho2023davidsonian} and TIFA~\cite{hu2023tifa}),
three human preference metrics
(Pick-a-Pic~\cite{kirstain2024pick}, ImageReward~\cite{xu2023imagereward}, and HPS~\cite{Wu2023HPS}),
and human evaluation.
Empirical results demonstrate that \methodname{} significantly improves T2I models' faithfulness to input text prompts and achieves higher human preference metrics.
Our final LoRA-Merging model achieves 6.9\% improvements on DSG, 2.1\% improvements on TIFA, and improves the human preference metrics by 0.4 on Pick-a-Pic, 0.39 on ImageReward, and 3.7 on HPS.
Furthermore, we empirically show that the
T2I models learned from the self-generated images achieve a performance similar to that of learning from ground-truth images (see \cref{fig:dataset}).
Lastly, we further show that fine-tuning with images from a weaker T2I model (\ie{}, SD v2) can help improve the faithfulness of a stronger T2I model (\ie{}, SDXL), suggesting promising weak-to-strong generalization in text-to-image models (see \Cref{table:weak_to_strong}). 

\section{Related Work}

\paragraph{Training Vision-Language Models with Synthetic Images.}
As recent denoising diffusion models~\cite{Sohl-Dickstein2015,Ho2020DDPM}
have achieved photorealistic image synthesis capabilities,
many works have studied using their synthetic images for training different models.
Azizi~\etal{}~\cite{Azizi2023},
Sariyildiz~\etal{}~\cite{Sariyildiz2023},
Lei~\etal{}~\cite{Lei2023},
inter alia,
study training image classification models with synthetic images.
For image captioning,
Caffagni~\etal{}~\cite{SynthCap2023} use diffusion models to generate images on the captioning data.
For training CLIP~\cite{Radford2021CLIP} models,
several works use diffusion models to generate images from existing captions~\cite{Tian2023StaleRep} or text generated with language models~\cite{Hammoud2024SynthCLIP}.
There is a recent research direction using synthetic images to train image generation models themselves, and we discuss more details in the following paragraph.

\paragraph{Training Text-to-Image Generation Models with Synthetic Images.}
A line of recent works train text-to-image (T2I) generation models
with synthetic images generated by the same or other models annotated with human preference scores using reinforcement learning~\cite{lee2023aligning,xu2023imagereward,Wu2023HPS,Dong2023RAFT,Clark2024DraFT,fan2023dpok} or direct preference optimization (DPO)~\cite{Rafailov2023DPO,wallace2023diffusion}.
While these works show promising results in improving model behavior with human preferences,
they require expensive human preference annotations.
SPIN-Diffusion~\cite{Yuan2024SPIN-Diffusion} proposes using self-play~\cite{SelfPlay1959,TD-gammon1995},
which was successfully adopted in Alphago Zero~\cite{alphagozero2016} and language models~\cite{Chen2024SPIN,Yuan2024Self-Reward-LM},
where the model itself becomes a judge and iteratively compares itself with previous iterations. However, self-play algorithm still relies on a set of ground truth image-text pairs as positive examples for supervision.
Concurrent/independent to our work,
DreamSync~\cite{sun2023dreamsync}
trains a T2I model by
first creating text prompts with LLMs,
sampling multiple images by the T2I model itself,
filtering out images with off-the-shelf scorers,
and fine-tuning the model on the resulting synthetic image-text pairs~\cite{sun2023dreamsync}.
Unlike DreamSync that depends on image filtering (generating 8 images and taking at most one of them for each text prompt,
\methodname{} generates 1 image for each prompt, significantly improving data generation efficiency by using only 2\% of image-text pairs compared with DreamSync.
Furthermore,
we focus on learning multiple skills with T2I models by 
learning and merging skill-specific LoRA experts
to mitigate knowledge interference across different skills, and we show this approach attains much stronger performance without adding any additional inference cost (see \Cref{table:lora_merging}). 

\section{\methodname{}: Learning and Merging
Text-to-Image
Skill-Specific
Experts
with Auto-Generated Data}
\label{sec:method}

\begin{figure}[t]
    \centering
    \includegraphics[width=0.95\textwidth]{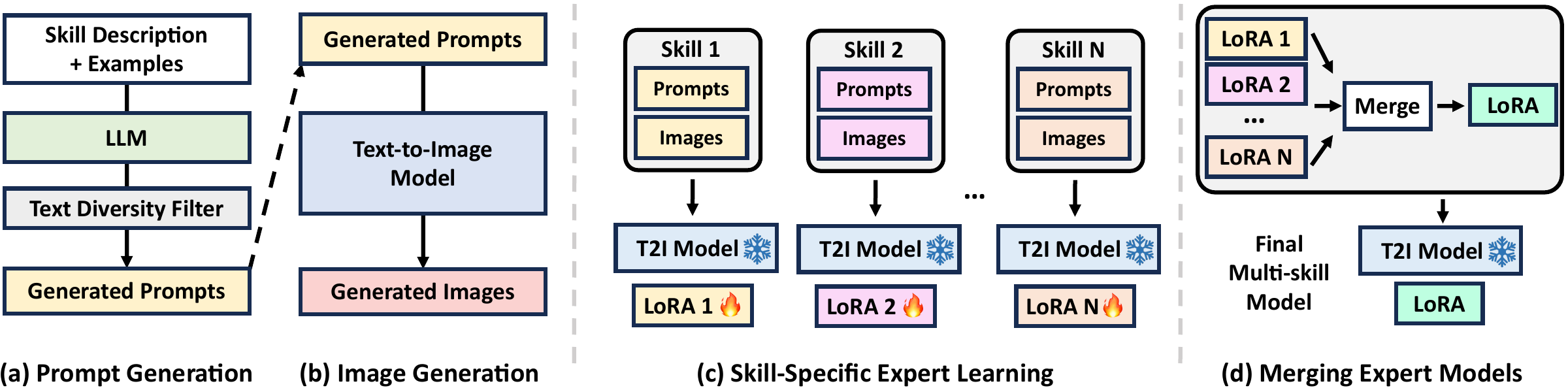}
    \caption{
    Illustration of the four-stage pipeline of \methodname{}.
    \textbf{(a) Prompt Generation}:
    Given a short skill description and a few (\ie{}, three) seed examples about a specific skill, we generate prompts to teach the skill with an LLM, while maintaining prompt diversity via text-similarity based filtering.
    \textbf{(b) Image Generation}:
    Given the LLM-generated text prompts, we
    generate training images with a T2I model.
    \textbf{(c) Skill-Specific Expert Learning}:
    We learn skill-specific expert T2I models based on LoRA fine-tuning.
    \textbf{(d) Merging Expert Models}:
    We obtain a multi-skill T2I model by merging the skill-specific LoRA parameters.
    }
   \label{fig:method}
\end{figure}

We introduce \methodname{},
a novel framework to
teach different skills to a T2I generation model
based on auto-generated data and model merging.
As illustrated in \cref{fig:method},
\methodname{} consists of four stages:
(1) skill-specific prompt generation with LLM (\cref{sec:sub:prompt_collect}),
(2) image generation with T2I Model (\cref{sec:sub:image_collect}),
(3) skill-specific expert learning (\cref{sec:sub:finetuning_lora}),
and
(4) merging expert models (\cref{sec:sub:lora_merging}).

\subsection{Automatic Skill-Specific Prompt Generation with LLM}
\label{sec:sub:prompt_collect}

As shown in \cref{fig:method} (a),
we automatically collect skill-specific prompts (that will be paired with images in \cref{sec:sub:image_collect})
to fine-tune T2I models in two steps:
(1) using large language models (LLMs) to generate prompts with brief skill descriptions and a few example prompts
and
(2) filtering the generated prompts to ensure their diversity.
In the following, we explain the two steps in detail.

\textbf{Prompt Generation.}
We leverage the in-context learning ability of LLMs to generate additional text prompts that follow similar writing styles (\eg{}, paragraph style) or acquire models' knowledge in the same domain (\eg{}, count capability). We manually collect three seed prompts with similar writing styles or acquire similar skills (\eg{}, spatial reasoning) to the target text prompts. 
Next, we use these seed prompts as in-context learning examples to query GPT-3.5 (\texttt{GPT3.5-turbo-instruct})~\cite{gpt35}.
We provide additional instructions that encourage diversity in generated prompts, including object occurrences, sentence patterns, and required skills for the T2I model to generate accurate prompts.
The detailed prompt template can be found in the Appendix.
During prompt generation,
we keep expanding the seed prompts with the generated prompts, and always randomly sample three prompts as in-context learning examples from the seed prompts. 

\textbf{Prompt Filtering.}
To improve the diversity of the collected text prompts, we filter out prompts that are similar to already generated ones.
As Taori~\etal{}~\cite{taori2023stanford} demonstrate that instruction diversity is crucial for improving the instruction following capability of large language models,
we follow the same intuition to create diverse text prompts. To ensure the diversity of generated prompts, we first receive a newly generated text prompt from the previous step. Then, we calculate its highest ROUGE-L~\cite{lin2004rouge} score with all the previously generated and filtered prompts.
Following Taori~\etal{}~\cite{taori2023stanford},
we discard text prompts with ROUGE-L$>$0.8 to maximize the diversity of generated prompts.

\subsection{Automatic Image Generation with Text-to-Image Models}
\label{sec:sub:image_collect}
As illustrated in~\cref{fig:method} (b), we generate corresponding images for each generated text prompt using the T2I model.
We find that existing diffusion-based T2I models are highly effective in learning from their self-generated images, and even benefit from learning from images generated with weaker T2I models (\Cref{table:weak_to_strong}).
It is important to leverage the knowledge
that already exists inside the T2I models
(learned from web data during pre-training),
and hence we aim to extract this knowledge
for creating the skill-specific image-text pairs, which in turn will be used to 
fine-tune the T2I models for improving faithfulness (\cref{sec:sub:finetuning_lora}).

\subsection{Fine-tuning with Multiple Skill-Specific LoRA Experts}
\label{sec:sub:finetuning_lora}

After constructing the self-training text-image pairs for different skills from previous steps, \methodname{} fine-tunes the T2I model on them to equip the model with these skills. We adopt Low-Rank Adaptation (LoRA)~\cite{hu2021lora} to efficiently adapt the model to different skills by learning skill-specific LoRA experts, which enables the T2I model to learn without having conflicts with the data in other skills~\cite{chen2024llavamole}.
LoRA also greatly reduces the fine-tuning cost when learning a large number of experts.
The LoRA fine-tuning optimizes the rank-decomposition matrices of dense layers in T2I models, making it more parameter-efficient compared to updating entire parameters of large T2I models.
Specifically, the updates to the weights $W_0 \in R^{d\times d}$ in the pre-trained T2I models can be represented as:
$W_0 + \Delta W = W_0 + BA$,
where $B \in R^{d\times r}$ and $A \in R^{r\times d}$ are low rank decomposition matrics. In practice, rank $r$ is selected to be much smaller (thus low-rank) than the hidden dimension $d$ for efficient fine-tuning. 

Fine-tuning T2I models on collected self-training data with skill-specific LoRA modules boosts T2I models' alignment on the specific text style or desired skills needed for faithfulness generation. Concretely, for each new dataset $t \in \mathcal{T}$, we fine-tune the T2I model with LoRA independently and this introduces $|\mathcal{T}|$ skill-specific LoRA modules after fine-tuning on all datasets (as shown in \cref{fig:method} (c)).
In \cref{sec:effectiveness_lora_merging}, we observe that learning and merging skill-specific experts is more effective than learning a single LoRA across all datasets,
by helping the T2I model mitigate knowledge conflicts between different skills~\cite{Liu2019LossBalancedTW}.

However, using multiple skill-specific experts requires the model to know which expert to use for a given input, and this usually requires user annotations on the skill category of inputs. An option is to learn a router to determine which expert to use in the test time~\cite{wu2023mole,chen2024llavamole}, but training router is inefficient as it requires the router to be trained on all datasets simultaneously or it may suffer from serious catastrophic forgetting~\cite{Kirkpatrick2016OvercomingCF}. In the next section, we propose to utilize model merging to efficiently construct a single final multi-skill model.

\subsection{Merging LoRA Expert Models to Obtain a Multi-Skill Model}
\label{sec:sub:lora_merging}

Recent work of model merging~\cite{Ilharco2022EditingMW,Singh2019ModelFV, Ainsworth2022GitRM,Sung2023AnEmpiricalSO,yadav2023tiesmerging} proposes methods that merge multiple task-specific weights into one, while mostly retaining the original task-specific performances.
Moreover, model merging can help mitigate the knowledge conflicts between datasets because we only need to adjust the merging ratios without re-training the task-specific models~\cite{Yu2023LanguageMA,Ram2023RewardedST}.
Due to these benefits, we then extend model merging to learn a final T2I model that can handle multiple skills without much knowledge conflicts.
Concretely, given $|\mathcal{T}|$ LoRA experts learned from \cref{sec:sub:finetuning_lora}, we merge all LoRA experts into one LoRA expert ($A^m = \frac{1}{|\mathcal{T}|}\sum_{t \in \mathcal{T}} A^t$ and $B^m = \frac{1}{|\mathcal{T}|}\sum_{t \in \mathcal{T}} B^t$) and this single expert can handle all skills simultaneously (as shown in \cref{fig:method} (d)).
With this approach, we can reach superior performance over standard multi-task LoRA training and even MoE-LoRA (learning a router with LoRA experts), as shown in \Cref{table:lora_merging,table:training_ablation}, and also eliminate the need to know the skill categories beforehand. Note that while ZipLoRA~\cite{Shah2023ZipLoRAAS} has demonstrated the use of LoRA merging (merging 2 LoRA modules) in diffusion models, to the best of our knowledge, we are the first to show the effectiveness of LoRA merging on multiple diverse skills (from 5 datasets) in diffusion models.

\section{Experimental Setup}
\label{sec:experiment_setup}

\subsection{Evaluation Benchmarks}
\label{sec:eval_benchmarks}

We evaluate models on two evaluation benchmarks that measure the alignment between text prompts and generated images:
\textbf{DSG}~\cite{cho2023davidsonian} and \textbf{TIFA}~\cite{hu2023tifa}.
Both benchmarks consist of prompts from different sources, covering diverse text prompt styles and generation skills.

\textbf{DSG}
consists of 1060 prompts from 10 different sources
(160 prompts from TIFA~\cite{hu2023tifa},
and 100 prompts from each of
{Localized Narratives}~\cite{pont2020connecting},
{DiffusionDB}~\cite{wang2022diffusiondb},
{CountBench}~\cite{paiss2023teaching},
{Whoops}~\cite{bitton2023breaking},
DrawText~\cite{liu2022character},
Midjourney~\cite{turc2023midjourney},
Stanford Paragraph~\cite{krause2017hierarchical},
VRD~\cite{lu2016visual},
PoseScript~\cite{delmas2022posescript}).
Among the ten DSG prompt sources,
we mainly experiment with text prompts from five prompt sources that have
(1) ground-truth image-text pairs (to compare the usefulness of auto-generated data with ground-truth data) and 
(2) measuring different skills required in T2I generation (\eg, following long captions, composing infrequent objects).
Specifically, we use 
\textbf{COCO}~\cite{lin2014microsoft} for short prompts with common objects in daily life,
\textbf{Localized Narratives}~\cite{pont2020connecting} for paragraph-style long captions, \textbf{DiffusionDB}~\cite{wang2022diffusiondb} for human-written prompts that specify many attribute details,
\textbf{CountBench}~\cite{paiss2023teaching} for evaluating object counting, and \textbf{Whoops}~\cite{bitton2023breaking} for commonsense-defying text prompts. 

\textbf{TIFA}
consists of 4,081 prompts from four sources, including
COCO~\cite{lin2014microsoft} for short prompts with common objects,
PartiPrompts~\cite{yu2022scaling} / DrawBench~\cite{saharia2022photorealistic} for challenging image generation skills,
and PaintSkills~\cite{Cho2023DallEval} for compositional visual reasoning skills.

\subsection{Evaluation Metrics}
\label{sec:eval_metrics}

We quantitatively evaluate the performance of T2I generation models in text faithfulness and human preference metrics.
See also \cref{sec:human_eval} for human evaluation.

\textbf{Text faithfulness.}
To evaluate T2I model's faithfulness in generation, we use VQA accuracy from TIFA and DSG. Specifically, TIFA and DSG utilize LLMs to generate questions given a text prompt and utilize the VQA model to check whether it can answer the questions correctly given the generated image. The image is considered to have better faithfulness to text prompts if the VQA model can answer the question more correctly.
For TIFA, 
we use BLIP-2 as the VQA model following Sun~\etal{}~\cite{sun2023dreamsync},
For DSG,
we use mPLUG-large~\cite{li2022mplug} as the VQA model,
as PaLI~\cite{chen2022pali} is not publicly accecssible, and Hu~\etal{}~\cite{hu2023tifa} shows that mPLUG achieves higher human correlation than BLIP-2.

\textbf{Human preference metrics.}
To evaluate how the generated images align with human preference, we use the PickScore~\cite{kirstain2024pick}, ImageReward~\cite{xu2023imagereward}, and HPS~\cite{Wu2023HPS}.
PickScore and HPS are based on CLIP~\cite{radford2021learning} trained on the Pick-a-Pic dataset~\cite{kirstain2024pick} and Human Preference Score dataset~\cite{Wu2023HPS} respectively, which both have annotations of human preference over images.
ImageReward is a BLIP~\cite{li2022blip} based reward model fine-tuned on human preference data collected on DiffusionDB.
We calculate PickScore, ImageReward, and HPS on the 1060 DSG prompts. We also provide the evaluation results on HPS prompts in the appendix.

\subsection{Implementation Details}
\label{sec:implemenation_details}

In the prompt generation stage (\cref{sec:sub:prompt_collect}),
we use \texttt{gpt-3.5-turbo-instruct}~\cite{gpt35} to generate text prompts by providing three prompts for each skill as in-context examples.
For each of the five datasets (COCO~\cite{lin2014microsoft},
{Localized Narratives}~\cite{pont2020connecting},
{DiffusionDB}~\cite{wang2022diffusiondb},
{CountBench}~\cite{paiss2023teaching}, and
{Whoops}~\cite{bitton2023breaking}),
we collect 1K prompts starting with three prompts randomly sampled from them, ensuring the prompts are not included in the DSG test prompts (\ie{}, 5K prompts in total).
We refer to the resulting auto-generated datasets as 
Localized Narrative$^{\text{\methodname{}}}$, CountBench$^{\text{\methodname{}}}$, DiffusionDB$^{\text{\methodname{}}}$, Whoops$^{\text{\methodname{}}}$, and COCO$^{\text{\methodname{}}}$.
We refer to the resulting combination of 5K auto-generated dataset as DSG$^{\text{\methodname{}-5K}}$.

In the image generation stage (\cref{sec:sub:image_collect}),
we use the default denoising steps 50 for all models, and the Classifier-Free Guidance (CFG)~\cite{ho2021classifierfree} of 7.5.
In the LoRA fine-tuning stage (\cref{sec:sub:finetuning_lora}), we use 128 as the LoRA rank.
We fine-tune LoRA in mixed precision (\ie{}, FP16) with a constant learning rate of 3e-4 and a batch size of 64. We fine-tune LoRA modules for 5000 steps, which is approximately 313 epochs. During inference, we uniformly merge the specialized LoRA experts into one multi-skill expert
(\cref{sec:sub:lora_merging}). 
We evaluate model checkpoints every 1000 steps and pick the model with the best text faithfulness on DSG benchmark. Fine-tuning LoRA for SD v1.4, SD v2, and SDXL takes 6 hours, 6 hours, and 12 hours on a single NVIDIA L40 GPU, respectively. 
We use Diffusers~\cite{von-platen-etal-2022-diffusers} for our experiments.

\section{Results and Analysis}
\label{sec:results}

We demonstrate the usefulness of \methodname{} with comprehensive experiments and analysis.
We first compare our proposed approach with multiple T2I methods that aim to improve the alignment of text and image (\cref{sec:comparison_models}).
Then we show the effectiveness of our proposed LoRA merging in mitigating knowledge conflict across different datasets (\cref{sec:effectiveness_lora_merging}),
and the effectiveness of auto-generated data by comparing it with ground truth data (\cref{sec:effectiveness_auto_data}).
Furthermore, we show
promising weak-to-strong generalization of T2I models (\cref{sec:weak_to_strong}),
human evaluation results (\cref{sec:human_eval}),
and ablation of training methods (\cref{sec:training_ablation}).
Lastly, we present 
qualitative examples (\cref{sec:qualitative_examples}).

\begin{table}[t]
  \caption{Comparison of \methodname{} and different 
  text-to-image alignment
  methods
  on text faithfulness and human preference (see \cref{sec:comparison_models} for discussion).
  \methodname{} achieves the best performance in all five metrics when adapted on different base models (\ie{}, SD v1.4, SD v2, and SDXL). Best scores for each model are in \textbf{bold}.
  }
  \label{table:main_quantitative}
  \resizebox{1.0\columnwidth}{!}{
  \centering
  \begin{tabular}{l l  cc  ccc}
    \toprule
   \multirow{2}{*}{\textbf{Base Model}}  & \multirow{2}{*}{\textbf{Methods}} & \multicolumn{2}{c}{\textbf{Text Faithfulness}} & \multicolumn{3}{c}{\textbf{Human Preference on DSG prompts}} \\ 
   \cmidrule(lr){3-4} \cmidrule(lr){5-7}
    &   & DSG$^{\text{mPLUG}}$ $\uparrow$ & TIFA$^{\text{BLIP2}}$ $\uparrow$ & PickScore $\uparrow$ & ImageReward $\uparrow$  & HPS $\uparrow$ \\
    \midrule
 \multirow{10}{*}{SD v1.4~\cite{rombach2022high}}& Base model  &  67.3 & 76.6 & 20.3 & -0.22  & 23.0 \\ 
 \cmidrule(lr){2-7}
 & \textit{(Training-free)}\\
 & SynGen~\cite{rassin2024linguistic} &  66.2 & 76.8  & 20.4 & -0.24 & 24.5 \\
&  StructureDiffusion~\cite{feng2022training} & 67.1 & 76.5 &  20.3 & -0.14  & 23.5 \\
\cmidrule(lr){2-7}
& \textit{(RL)} \\
 & DPOK~\cite{fan2023dpok} & - & 76.4  & - & -0.26  & - \\
 & DDPO~\cite{black2023training} & - & 76.7   & - & -0.08  & - \\ 
 \cmidrule(lr){2-7}
 & \textit{(Automatic data generation)} \\
 & DreamSync~\cite{sun2023dreamsync} & - & 77.6 &  - & -0.05 & - \\
 & \textbf{\methodname{} (Ours)} & \textbf{71.3} & \textbf{79.5} &  \textbf{20.5} & \textbf{0.36} & \textbf{25.5}  \\
 \midrule 
 \multirow{2}{*}{SD v2~\cite{rombach2022high}} & Base model & 70.3 & 79.2 &  20.8 & 0.17  & 24.0 \\
 & \textbf{\methodname{} (Ours)} & \textbf{77.7} & \textbf{83.2} &  \textbf{21.3} & \textbf{0.72} & \textbf{27.5} \\
 \midrule 
  \multirow{3}{*}{SDXL~\cite{podell2023sdxl}} & Base model & 73.3 & 83.5 &  21.6 & 0.70  & 26.2 \\
  & DreamSync~\cite{sun2023dreamsync} &  - & 85.2  & - & 0.84  & - \\
  & \textbf{\methodname{} (Ours)} & \textbf{80.2} & \textbf{85.6} &  \textbf{22.0} & \textbf{1.09} & \textbf{29.9} \\
    \bottomrule
  \end{tabular}
  }
\end{table}

\subsection{Comparison with
Different Alignment Methods for Text-to-Image Generation
}
\label{sec:comparison_models}

We compare \methodname{} with different alignment methods for T2I generation, including training-free methods (SynGen~\cite{rassin2024linguistic}, StructureDiffusion~\cite{feng2022training}),
RL-based methods (DPOK~\cite{fan2023dpok}, DDPO~\cite{black2023training}),
and DreamSync~\cite{sun2023dreamsync}, a concurrent method based on automatic data generation.
We experiment with three diffusion-based T2I models (\ie{}, SD v1.4, SD v2, and SDXL).

\textbf{\methodname{} outperforms other alignment methods for T2I generation.}
As shown in \Cref{table:main_quantitative}, \methodname{} consistently improves faithfulness and human preference metrics for all three backbones.
Specifically, on SD v1.4, \methodname{} improves the baseline by \textbf{2.9\%} in TIFA, \textbf{4.0\%} in DSG, \textbf{0.2} in PickScore, \textbf{0.58} in ImageReward, and \textbf{2.5} in HPS score.
Furthermore, \methodname{}
achieves significantly higher performance than other baselines, including the RL-based methods (DPOK/DDPO), which require annotated human preference data,
and DreamSync, a concurrent/independent work based on a larger auto-generated dataset (\ie{}, 28K text prompts; \methodname{} uses 5K text training prompts in total),
and image filtering (\ie{}, generating 8 images and taking at most one of them for each text prompt; \methodname{} only generates 1 image for each prompt).
Besides, on SD v2 and SDXL, \methodname{} shows larger improvement in text faithfulness (\ie{}, \textbf{7.4\%} improvement on DSG for SD v2, and \textbf{6.9\%} on DSG for SDXL),
demonstrating the effectiveness of \methodname{}.

\begin{table}[t]
  \caption{
  Comparison of single LoRA and LoRA Merging in text faithfulness and human preference
  (see \cref{sec:effectiveness_lora_merging} for discussion).
  We use SD v2 as our base model and train models with our automatically generated image-text pairs.
  \texttt{DATA}$^{\text{\methodname{}}}$: auto-generated image-text pairs where
  prompts are generated with LLMs with three prompt examples from \texttt{DATA} that are not included in DSG test prompts (see \cref{sec:implemenation_details} for details).
  \textit{LN: Localized Narratives;
  CB: CountBench;
  DDB: DiffusionDB.}
  Best/2nd best scores are \textbf{bolded}/\underline{underlined}.
  }
  \label{table:lora_merging}  
  \resizebox{1.0\columnwidth}{!}{
  \centering
  \begin{tabular}{l l ccccc cc ccc}
    \toprule
 \multirow{4}{*}{\textbf{No.}} & \multirow{4}{*}{\textbf{Model}} & \multicolumn{5}{c}{\textbf{Auto-Generated Training Dataset}}   & \multicolumn{2}{c}{\textbf{Text Faithfulness}} & \multicolumn{3}{c}{\textbf{Human Preference on DSG}} \\ 
   \cmidrule(lr){3-7} \cmidrule(lr){8-9} \cmidrule(lr){10-12}
    & &  LN$^{\text{\methodname{}}}$     & CB$^{\text{\methodname{}}}$ & DDB$^{\text{\methodname{}}}$ & \multirow{1}{*}{Whoops$^{\text{\methodname{}}}$} &  COCO$^{\text{\methodname{}}}$ & \multirow{3}{*}{DSG$^{\text{mPLUG}}$} & \multirow{3}{*}{TIFA$^{\text{BLIP2}}$} & \multirow{3}{*}{PickScore} &  \multirow{3}{*}{ImageReward}   & \multirow{3}{*}{HPS}  \\
    & & \multirow{2}{*}{\textit{(Paragraph)}}  & \multirow{2}{*}{\textit{(Count)}} & \textit{(Real} &  \textit{(Counter-} & \textit{(Common}& &  &  &   &   \\
    & &  &  &  \textit{Users)} & \textit{Factual)} & \textit{Objects)} && \\
    \midrule
0. & SDv2 & &&&&&70.3 & 79.2 & 20.8 & 0.17 & 24.0  \\ 
\midrule
1. & \multirow{7}{*}{+ Single LoRA} & \cmark &&&& &76.4	& 81.4  & 20.9 & 0.56 & 26.2 \\
2.  &  & & \cmark &&&& 76.0 & 81.4 & 20.8 & 0.46 & 25.7	\\
3. &  &  && \cmark&&& 73.0 &	81.2 & 20.9 & 0.46 & 25.8 \\
4. &  &  &&& \cmark&& 73.0 & 80.7 &20.8 & 0.44 & 25.3 \\ 
5. &  &  &&&& \cmark& 76.0 & 81.3 & 20.9 & 0.47 & 25.6 	\\
6. &  & \cmark&\cmark&\cmark&&& 75.1	& 81.5 & 20.7 & 0.37 & 24.8 \\
7. &  & \cmark&\cmark&\cmark&\cmark&\cmark& 74.4 & 80.2 & 20.6 & 0.35 & 24.9 \\
\midrule
8. & \multirow{2}{*}{+ LoRA Merging} & \cmark&\cmark&\cmark&&& \underline{76.9}	& \underline{82.9} & \underline{21.2} & \underline{0.65} & \underline{27.3} \\ 
9. &  & \cmark&\cmark&\cmark&\cmark&\cmark& \textbf{77.7}	& \textbf{83.2} & \textbf{21.3} & \textbf{0.72} & \textbf{27.5} \\ 
    \bottomrule
  \end{tabular}
  }
\end{table}

\subsection{Effectiveness of Learning \& Merging Skill-Specific Experts}
\label{sec:effectiveness_lora_merging}
We compare
(1) separately learning multiple LoRA experts on different auto-generated datasets followed by merging
and 
(2) training a single LoRA on a mixture of datasets.
For this, 
we experiment with our five auto-generated image-text pairs:
Localized Narrative$^{\text{\methodname{}}}$, CountBench$^{\text{\methodname{}}}$, DiffusionDB$^{\text{\methodname{}}}$, Whoops$^{\text{\methodname{}}}$, and COCO$^{\text{\methodname{}}}$ (see \cref{sec:implemenation_details} for details).

\textbf{Learning and merging skill-specific LoRA experts is more effective than single LoRA on multiple datasets.}
As shown in \Cref{table:lora_merging},
the LoRA models trained separately on each of the five automatically generated datasets
(\textit{No.1.} to \textit{No.5.}) can improve the overall metric over the baseline SD v2 -- 70.3\%, while the degree of improvements is different for each metric
(\eg{}, 76.4\% for fine-tuning with Localized Narrative$^{\text{\methodname{}}}$, and 73.0\% for fine-tuning with DiffusionDB$^{\text{\methodname{}}}$).
However, training multiple skills simultaneously with a single LoRA (\textit{No.6.} to \textit{No.7.}) tends to degrade performance as more datasets are incorporated. This indicates that the T2I model struggles with LoRA to accommodate distinct skills and writing styles from different datasets. A similar phenomenon has been reported in LLaVA-MoLE~\cite{chen2024llavamole}, where the knowledge conflict between datasets can degrade the performance of multi-task training. We find that merging multiple skill-specific LoRA experts (\textit{No.8.} and \textit{No.9.}) achieves the best performance in both text faithfulness and human preference, demonstrating that merging LoRA experts can help mitigate the knowledge conflict between multiple skills.

\begin{figure}[t]
    \begin{center}
    \includegraphics[width=.75\textwidth]{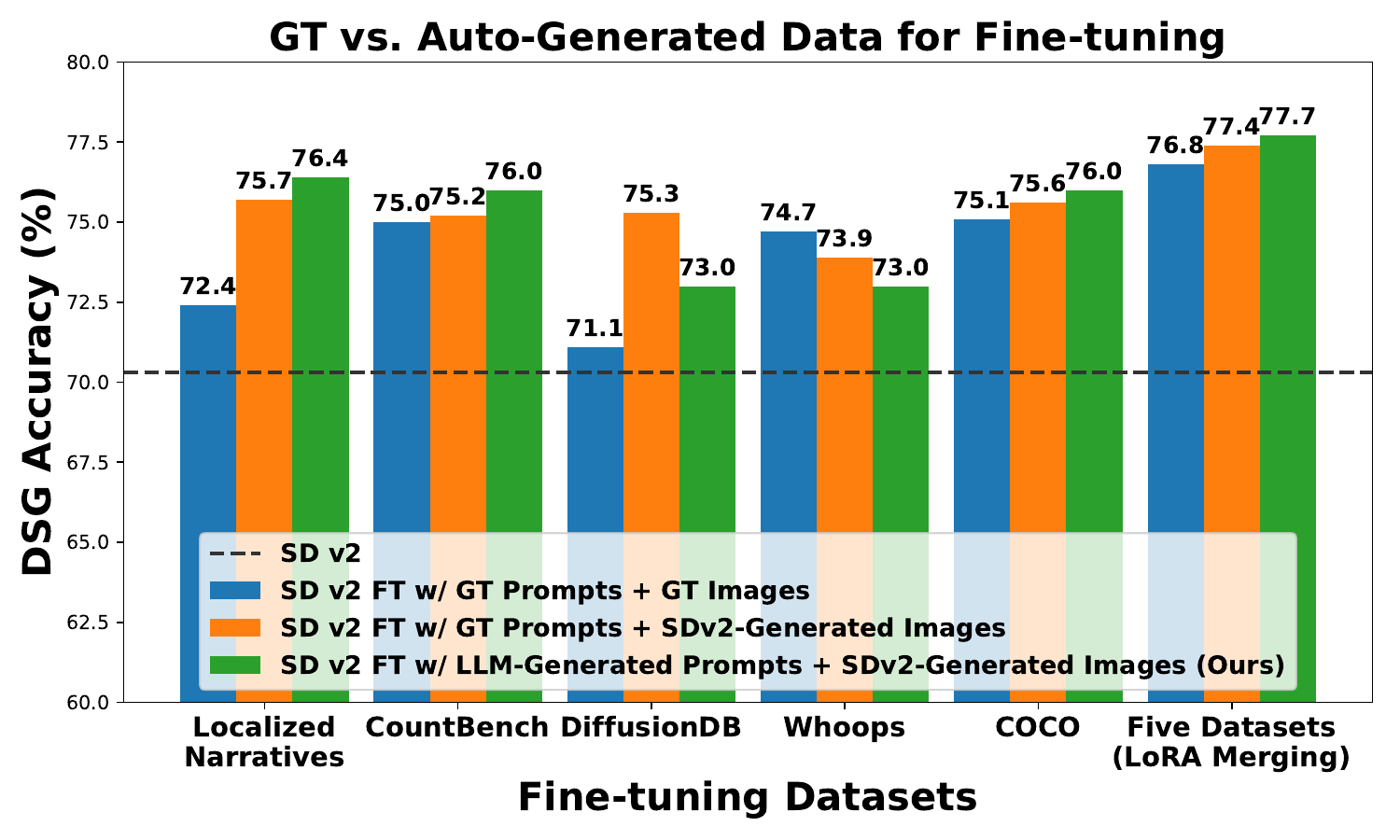}
    \end{center}
    \caption{
    DSG accuracy of SD v2 fine-tuned with different image-text pairs.
    }
   \label{fig:dataset}
\end{figure}

\subsection{Effectiveness of Auto-Generated Data}
\label{sec:effectiveness_auto_data}

In this section, we investigate the effectiveness of our automatically generated data by comparing them with ground truth data. We fine-tune SD v2 model using ground truth data from Localized Narratives, CountBench, DiffusionDB, Whoops, and COCO, sampling 1K image-text pairs from each dataset and fine-tuning specialized LoRA experts accordingly.

\textbf{Fine-tuning with auto-generated data can achieve comparable performance to fine-tuning with ground truth data.}
As shown in \cref{fig:dataset}, we observe that fine-tuning with either auto-generated or ground truth data improves from baseline SD v2 performance -- 70.3\%, when evaluated on the DSG benchmark.
Surprisingly, fine-tuning with the generated data via  \methodname{} outperforms the use of ground truth data in most cases, leading to a DSG accuracy improvement of \textbf{4.0\%} with Localized Narrative style prompts, \textbf{1.0\%} with CountBench style prompts, \textbf{1.9\%} with DiffusionDB style prompts, and \textbf{0.9\%} with COCO style prompts.
In short, our approach results in an average improvement of 1.2\% brought by fine-tuning only auto-generated data without any need for human-collected ground truth text-image pairs,
suggesting that diffusion-based text-to-image models may benefit from the diversity of self-generated images. Furthermore, we investigate whether the improvement is brought by text prompt or image quality. We generate images with SD v2 based on 1K ground truth captions, and fine-tune specialized LoRA experts accordingly. We observe that in most of the cases, using generated images works better than ground truth images (\eg{}, Localized Narrative), suggesting T2I models can generate images with comparable alignment as ground truth images. Besides, learning from our LLM-generated captions achieves comparable performance with learning from ground truth captions, suggesting the effectiveness of our text prompt collection process.

\begin{table}
  \caption{Comparison of different image generators for creating training images.
  In addition to using the same model being trained as an image generator, we also experiment with using a smaller model as an image generator (No. 4.).
  SDXL is a bigger and stronger model than SD v2.
  See \cref{sec:weak_to_strong} for discussion.
  }
  \label{table:weak_to_strong}
  \resizebox{1.0\columnwidth}{!}{
  \centering
  \begin{tabular}{l c c cc ccc}
    \toprule
  \multirow{2}{*}{\textbf{No.}} & \multirow{2}{*}{\textbf{Base Model}}  & \multirow{2}{*}{\textbf{Training Image Generator}} & \multicolumn{2}{c}{\textbf{Text Faithfulness}} & \multicolumn{3}{c}{\textbf{Human Preference on DSG}} \\ 
   \cmidrule(lr){4-5} \cmidrule(lr){6-8}
   & & & DSG$^{\text{mPLUG}}$ $\uparrow$ & TIFA$^{\text{BLIP2}}$ $\uparrow$ & PickScore $\uparrow$  & ImageReward $\uparrow$  & HPS $\uparrow$  \\
   \midrule 
  1. & SD v2 & - & 70.3 & 79.2 & 20.8 & 0.17 & 24.0 \\
  2. & SDXL & - & 73.3 & 83.5 & 21.6 & 0.70 & 26.2 \\ 
  \midrule
  3. &SD v2 & SD v2 & 77.7 & 83.2 & 21.3 & 0.72  & 27.5  \\ 
  4. &SDXL & SD v2 & \textbf{81.3} & 83.8 & 21.5 & 0.78 & 28.8 \\ 
  5. & SDXL & SDXL & 80.2 & \textbf{85.6} & \textbf{22.0} & \textbf{1.09} & \textbf{29.9} \\
    \bottomrule
  \end{tabular}

  }
\end{table}

\subsection{Weak-to-Strong Generalization} 
\label{sec:weak_to_strong}

In previous experiments, we demonstrate the interesting self-improving capabilities of
T2I models,
where
the training images were generated by the same T2I model.
Here, we delve into the following research question: 
\textit{``Can a T2I model benefit from learning with images generated by a weaker model?''}.
The problem of \textit{weak-to-strong} generalization was initially explored in the context of LLMs~\cite{burns2023weak, Saha2023CanLM},
referred to as superalignment,
which involved training GPT-4~\cite{Achiam2023GPT4TR} using responses generated by a weaker agent, such as GPT-2.

\textbf{Weaker T2I models can help stronger T2I models.}
As shown in \Cref{table:weak_to_strong}, fine-tuning SDXL with generated images from SD v2 (\textit{No.4.}) remarkably enhances performance over the SDXL baseline (\textit{No.2.}) in both text faithfulness and human preference.
In addition, this approach achieves competitive performance compared with fine-tuning SDXL with SDXL-generated images (\textit{No.5.}), indicating a promising potential for weak-to-strong generalization in diffusion-based T2I generation models.
To the best of our knowledge, this is the first work to find promising improvements in the weak-to-strong generalization for text-to-image diffusion models.

\begin{figure}[t]
    \centering
    \includegraphics[width=\textwidth]{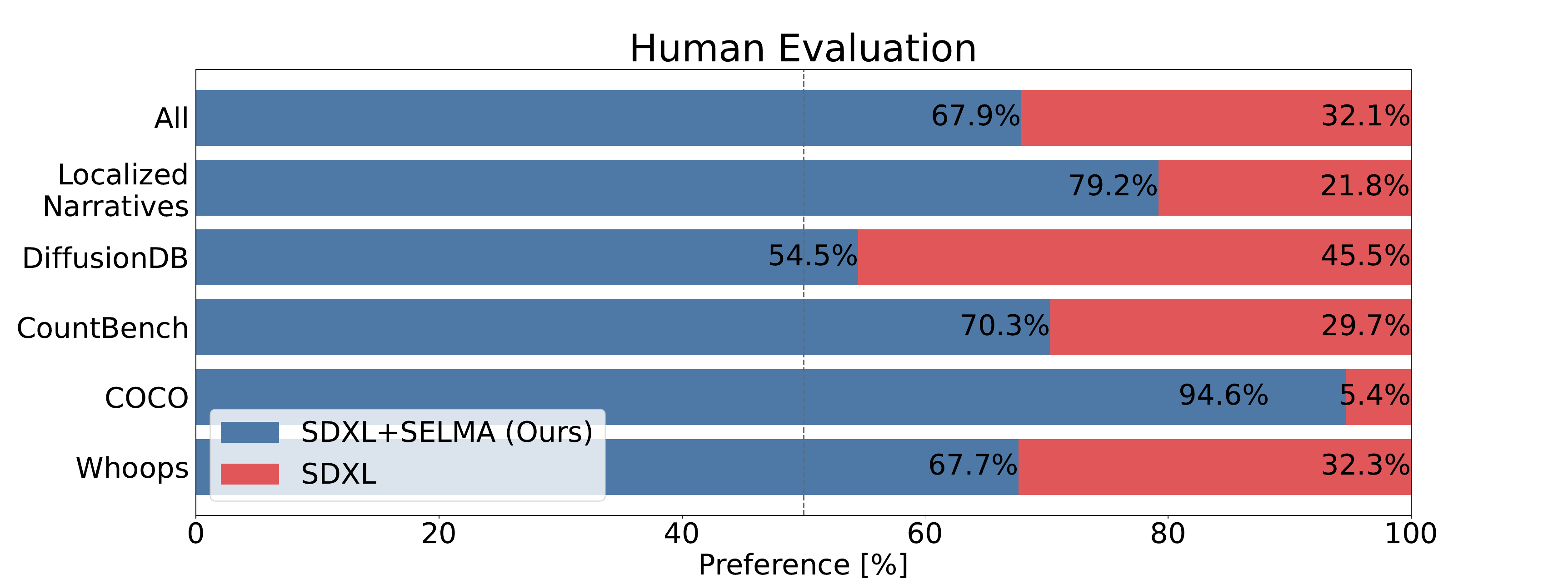}
    \caption{Human Evaluation on 200 sampled text prompts from DSG,
    where we
    show the win \vs lose percentages
    of SDXL and SDXL+\methodname{} (Ours).
    }
   \label{fig:human_eval}
\end{figure}

\subsection{Human Evaluation}
\label{sec:human_eval}

In addition to automatic evaluation using text faithfulness benchmarks (DSG and TIFA) and human preference metrics (PickScore, ImageReward, and HPS), we further perform a human evaluation to compare the performance of SDXL and SDXL fine-tuned with \methodname{} on DSG$^{\text{\methodname{}-5K}}$ (details in \cref{sec:implemenation_details}).
We randomly select 200 prompts from DSG and ask three annotators to determine ``Which image aligns with the caption better?'' given the text prompt and generated images from both SDXL and SDXL+\methodname{}.
We provide win/tie/lose options to the annotators, and we report the win \vs{} lose percentage in the following.
The user interface, instructions, and the detailed statistics are provided in appendix.

\textbf{SDXL+\methodname{} 
is preferred than SDXL in terms of text alignment.} 
\cref{fig:human_eval} shows that
on all five DSG splits, images generated with SDXL+\methodname{} are preferred \textbf{67.9\%} of the time, compared to 32.1\% for the baseline SDXL. Furthermore, on the five datasets fine-tuned with similar text prompts,  SDXL+\methodname{} achieve a preference rate of \textbf{94.6\%} on COCO split and \textbf{79.2\%} on Localized Narratives. This substantial preference over the baseline model demonstrates the effectiveness of  \methodname{} in enhancing T2I models.

\begin{table}
  \caption{Comparison with different fine-tuning methods on SD v2 with our auto-generated data,
  in text faithfulness and human preference. See \cref{sec:training_ablation} for discussion.
  }
  \label{table:training_ablation}
  \resizebox{1.0\columnwidth}{!}{
  \centering
  \begin{tabular}{l l cc ccc}
    \toprule
   \multirow{2}{*}{\textbf{No.}}   & \multirow{2}{*}{\textbf{Methods}} & \multicolumn{2}{c}{\textbf{Text Faithfulness}} & \multicolumn{3}{c}{\textbf{Human Preference on DSG}} \\ 
   \cmidrule(lr){3-4} \cmidrule(lr){5-7}
   & & DSG$^{\text{mPLUG}}$ $\uparrow$ & TIFA$^{\text{BLIP2}}$ $\uparrow$ & PickScore $\uparrow$  & ImageReward $\uparrow$ & HPS $\uparrow$ \\
   \midrule 
   0. & SDv2 &  70.3 & 79.2  & 20.8 & 0.17  & 24.0 \\
   1. & + LoRA Merging (\methodname{}) & \textbf{77.7} & \textbf{83.2} & \textbf{21.3} & \textbf{0.72}  & \textbf{27.5} \\ 
   2. & + LoRA Merging + DPO & 75.1 & 81.4 & 20.8 & 0.44 & 26.0 \\
   3. & + MoE-LoRA & 77.2 & 83.0 & \textbf{21.3} &  0.68 & 27.2 \\ 
    \bottomrule
  \end{tabular}
  }
\end{table}

\begin{figure}[t]
    \centering
    \includegraphics[width=.95\textwidth]{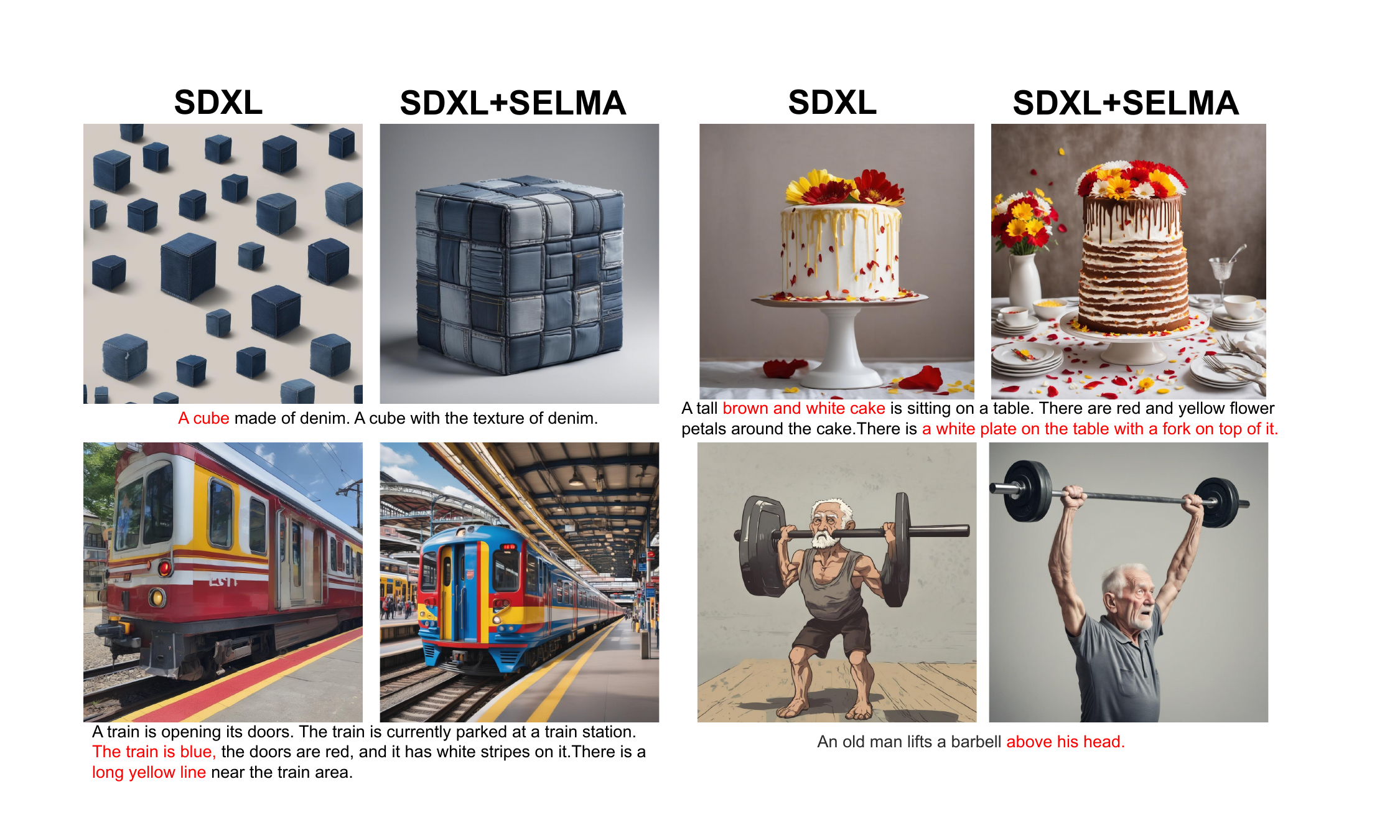}
    \caption{Example images generated with SDXL and SDXL+\methodname{}.
    \methodname{} shows better performance in object composition, attribute binding, and long text prompt following.
    We highlight the parts of the prompts in \textcolor{red}{red} where SDXL makes errors while  SDXL+\methodname{} generates correctly.
    }
   \label{fig:qual_main}
\end{figure}

\subsection{Training Method Ablations}
\label{sec:training_ablation}

We experiment with various training configurations for \methodname{} to validate our design choices for fine-tuning.
As our current experiments are based on 
supervised fine-tuning with LoRA Merging,
we additionally explore Direct Preference Optimization (DPO)~\cite{Rafailov2023DPO, wallace2023diffusion} as an alternative to supervised fine-tuning and employing Mixture of Lora Experts (MoE-LoRA)~\cite{wu2023mole} instead of LoRA Merging.
See the Appendix for the implementation details. 
\Cref{table:training_ablation} demonstrates that while fine-tuning with DPO and MoE-LoRA significantly improves the T2I models' text faithfulness and human preference (\textit{No.2 \& 3.} \vs{} \textit{No.0.}), simple inference-time LoRA merging achieves the best overall performance. In the end, we adopt LoRA merging and supervised fine-tuning as the default configuration in \methodname{} for its simplicity and efficiency.

\subsection{Qualitative Examples}
\label{sec:qualitative_examples}

We show some qualitative examples of images generated with SDXL fine-tuned with  \methodname{} paradigm in \cref{fig:qual_main}. We find that fine-tuning with  \methodname{} improves SDXL's capability in composing infrequently co-occurred attributes (\ie{}, ``cube'' and ``denim'' in the top-left image), composing multiple objects mentioned in the text prompts (\ie{}, ``brown and white cake'', ``table'', ``red and yellow flower'', and ``fork'' in the top-right image), following details in the long paragraph-style text prompts (\ie{}, ``blue train with white stripes'' and ``long yellow line near train area'' in the bottom-left image), and generating images that challenge commonsense (\ie{}, ``Old man lifts a barbell'' in bottom-right image). These qualitative examples demonstrate the effectiveness of  \methodname{} in improving T2I models' text faithfulness and human preference.

\section{Conclusion}
\label{sec:conclusion}

We propose \methodname{}, an novel paradigm to improve state-of-the-art T2I models' faithfulness in generation and human preference by
eliciting the pre-trained knowledge
of
T2I models.
\methodname{} first collects self-generated images given diverse generated text prompts without the need for additional human annotation.
Then, \methodname{} fine-tunes separate LoRA models on different datasets and merges them during inference to mitigate knowledge conflict between datasets.
\methodname{} demonstrates strong empirical results in improving T2I models' faithfulness and alignments to human preference and suggests potential weak-to-strong generalization for diffusion-based T2I models. 

\section{Acknowledgement}
We thank Elias Stengel-Eskin and Prateek Yadav for the thoughtful discussion. This work was supported by DARPA ECOLE Program No. HR00112390060, NSF-AI Engage Institute DRL-2112635, DARPA Machine Commonsense (MCS) Grant N66001-19-2-4031, ARO Award W911NF2110220, ONR Grant N00014-23-1-2356, and a Bloomberg Data Science Ph.D. Fellowship. The views contained in this article are those
of the authors and not of the funding agency.

\medskip

\bibliographystyle{splncs04}
\bibliography{main}

\appendix

\include{appendix}

\end{document}

%% file: appendix.tex
\section*{Appendix}
In this appendix, we present the following:
\begin{itemize}
    \item Evaluation on HPS v2.1 benchmark (\cref{sec:hps}).
    \item Additional qualitative examples with \methodname{} on SDXL backbone (\cref{sec:qual_appendix}).
    \item Skill-specific VQA accuracy on both TIFA and DSG benchmarks (\cref{sec:skill-specific-acc})
    \item Human evaluation details (\cref{sec:detail_human_eval}). 
    \item Implementation details of two training configuration variants: DPO and MoE-LoRA (\cref{sec:dpo_moelora_imp}).
    \item Prompts we used to query LLM to generate new text data (\cref{sec:prompt_detail}).
    \item Limitations of \methodname{} approach (\cref{sec:limitation}).
\end{itemize}

\section{Evaluation on HPS v2.1 Benchmark} \label{sec:hps}
In the main paper, we calculate HPS score~\cite{Wu2023HPS} on text prompts on DSG benchmark, following DreamSync~\cite{sun2023dreamsync}.
In this section, we additionally show the HPS score on the prompts from HPS v2.1 benchmark. HPS benchmark contains 3200 unique prompts from four different categories: anime, concept-art, paintings, and photo. We calculate the HPS score based on its HPS v2.1 model trained on higher quality datasets. As shown in \Cref{table:hps_eval}, when adapting \methodname{} to different stable diffusion base model, our approach significantly improves the baseline performance (\ie{} 2.5 for SD v1.4, 4.3 for SD v2, and 1.4 for SDXL), achieving better performance than all the released model on the HPS benchmark\footnote{Benchmark performance can be found: \url{https://github.com/tgxs002/HPSv2}}.

\begin{table}[h]
  \caption{Evaluation on HPS v2.1 evaluation benchmark. \methodname{} achieves signifiantly better scores on HPS evaluation benchmark compared with baselines, and 
  outperforming all other baselines reported in
  HPS v2.1 benchmark.
  }
  \label{table:hps_eval}
  \centering
  \begin{tabular}{l ccccc}
    \toprule
   \multirow{2}{*}{\textbf{Method}} &\multicolumn{5} {c}{\textbf{HPS v2.1 Evaluation Benchmark}}  \\ 
   \cmidrule(lr){2-6}
   & Anime & Concept-Art & Paintings & Photo & Average \\ 

   \midrule 
   SD v1.4~\cite{rombach2022high} &  26.0	& 24.9	& 24.8	& 25.7	& 25.4 \\ 
   + \methodname{} & \textbf{28.2} & \textbf{27.6} & \textbf{27.8} & \textbf{28.0} & \textbf{27.9} \\ 
   \midrule 
    SD v2~\cite{rombach2022high} & 27.1	& 26.0	& 25.7	& 26.7	& 26.4 \\ 
    + \methodname{}  &  \textbf{32.0} & \textbf{30.3} & \textbf{30.0} & \textbf{30.4} & \textbf{30.7} \\
    \midrule
    SDXL~\cite{podell2023sdxl} & 33.3 &	32.1 &	31.6 & 	28.4 & 	31.3 \\ 
    + \methodname{} & \textbf{34.7} & \textbf{32.7} & \textbf{32.6} & \textbf{30.8} & \textbf{32.7} \\
    \bottomrule
  \end{tabular}
\end{table}

\begin{figure}[]
    \centering
    \includegraphics[width=.85\textwidth]{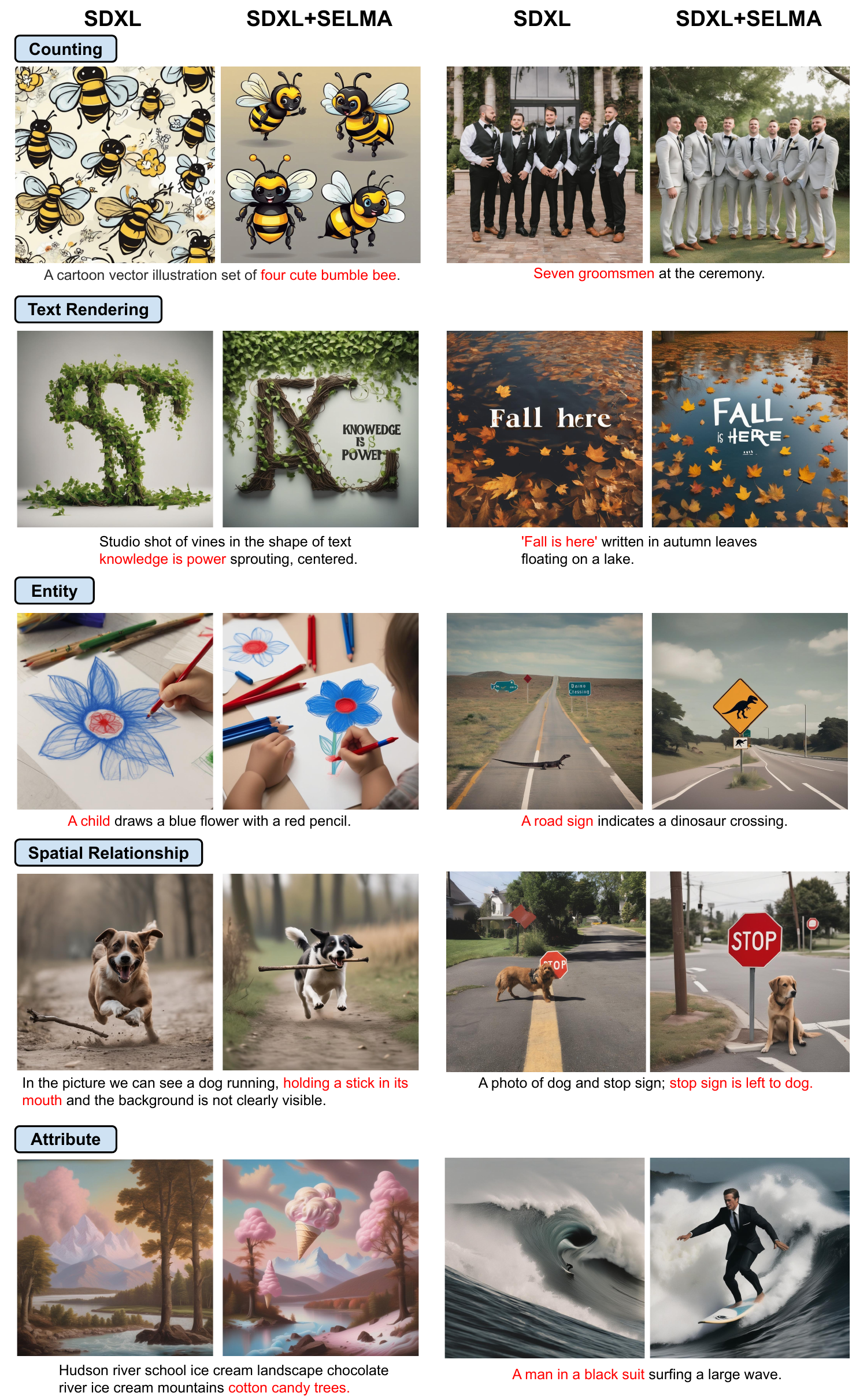}
    \caption{Qualitative example images generated with SDXL and SDXL+\methodname{} (Ours) from DSG~\cite{cho2023davidsonian} test prompts requiring different skills.
    \methodname{} helps improve SDXL in various skills, including counting, text rendering, spatial relationships, and attribute binding.
    We highlight the parts of the prompts in \textcolor{red}{red} where SDXL makes errors while  SDXL+\methodname{} generates correctly.
    }
   \label{fig:qual_appendix}
\end{figure}

\section{Additional Qualitative Exmaples} \label{sec:qual_appendix}

In \cref{fig:qual_appendix}, we show additional qualitative examples of SDXL and SDXL+\methodname{}  from DSG~\cite{cho2023davidsonian} test prompts requiring different skills.
\methodname{} helps improve SDXL in various skills, including counting, text rendering, spatial relationships, and attribute binding.
For counting skill prompts,
SDXL+\methodname{} generates ``four bees'' and ``seven groomsmen'' correctly following the text prompts.
For text rendering skill prompts,
SDXL+\methodname{} can render the text (``knowledge is power'' and ``Fall is here'') more accurately, while
it still lacks the capability to render the text in the texture of vines or autumn leaves.
For entity skill (placing correct objects) prompts,
the SDXL sometimes
 misses some entities
mentioned in the text prompt (\ie{}, ``A child'', and ``A road sign''), while SDXL+\methodname{} can successfully generate them.
For spatial relationship skill prompts,
SDXL+\methodname{} generated images (\ie{}, ``holding a stick in its mouth'', and ``stop sign left to dog'').
Lastly, for attribute skill prompts,
SDXL+\methodname{} binds objects with their corresponding attributes (\ie{}, ``cotton candy trees'' and ``A man in black suit'')  
more accurately than SDXL.
These qualitative results demonstrate the effectiveness of \methodname{}.

\begin{table}[t]
  \caption{Detailed skill-specific comparison of 
  SD models \vs{} SD models+\methodname{}
  on TIFA benchmark. 
  }
  \label{table:tifa_skill}
  \resizebox{1.0\columnwidth}{!}{
  \centering
  \begin{tabular}{l ccccccccccccc}
    \toprule
   \multirow{2}{*}{\textbf{Method}} &\multicolumn{13} {c}{\textbf{TIFA skills}}  \\ 
   \cmidrule(lr){2-14}
   & Animal/Human & Object & Location & Activity & Color & Spatial & Attribute & Food & Counting & Material & Other & Shape & Average \\ 
   \midrule 
   SD v1.4~\cite{rombach2022high} & 83.7 & 78.3 & 80.3 & 71.7 & 73.0 & 58.9 & 74.5 & 81.8 & 63.3 & 76.6 & 47.3 & \textbf{65.2} & 75.8 \\ 
   + \methodname{} & \textbf{87.1} & \textbf{83.0} & \textbf{84.8} & \textbf{75.9} & \textbf{74.4} & \textbf{62.3} & \textbf{76.0} & \textbf{88.1} & \textbf{66.2} & \textbf{78.5} & \textbf{52.2} & 59.4 & \textbf{79.5} \\ 
   \midrule 
SD v2~\cite{rombach2022high} & 86.5 & 82.6 & 83.8 & 75.6 & 76.8 & 62.4 & 75.4 & 85.2 & \textbf{66.5} & \textbf{82.2} & 55.4 & \textbf{75.0} & 79.2 \\ 
+ \methodname{}  & \textbf{89.7} & \textbf{88.0} & \textbf{87.6} & \textbf{80.3} & \textbf{80.3} & \textbf{66.0} & \textbf{77.2} & \textbf{91.0} & 65.8 & 81.3 & \textbf{63.2} & 68.1 & \textbf{83.2}  \\
    \midrule
    SDXL~\cite{podell2023sdxl} & 90.3 & 86.4 & 86.6 & 80.0 & 78.6 & 67.7 & 78.3 & 90.6 & 67.4 & \textbf{84.2} & 67.7 & \textbf{62.3} & 82.9 \\ 
    + \methodname{} & \textbf{93.4} & \textbf{90.4} & \textbf{89.5} & \textbf{83.6} & \textbf{81.1} & \textbf{69.6} & \textbf{78.5} & \textbf{92.1} & \textbf{68.8} & 83.7 & \textbf{68.7} & 60.9 & \textbf{85.6} \\
    \bottomrule
  \end{tabular}
  }
\end{table}

\begin{table}[t]
  \caption{Detailed skill-specific comparison of 
  SD models \vs{} SD models+\methodname{}
  on DSG benchmark.
  We show the skill categories that have more than 50 questions.
  }
  \label{table:dsg_skill}
  \resizebox{1.0\columnwidth}{!}{
  \centering
  \begin{tabular}{l ccccccccccccc}
    \toprule
   \multirow{2}{*}{\textbf{Method}} &\multicolumn{13} {c}{\textbf{DSG skills}}  \\ 
   \cmidrule(lr){2-14}
   & Whole & Color & Shape & Spatial & Part & State & Count & Action & Global & Material & Type & Text Rendering & Average \\ 
   \midrule 
   SD v1.4~\cite{rombach2022high} & 78.6 & \textbf{62.5} & 46.0 & 61.1 & 68.1 & 58.2 & 62.4 & 59.9 & \textbf{59.4} & 42.3 & \textbf{73.0} & \textbf{52.7} & 67.2\\ 
   + \methodname{} & \textbf{83.7} & \textbf{62.5} & \textbf{52.0} & \textbf{66.2} & \textbf{72.1} & \textbf{63.3} & \textbf{66.1} & \textbf{72.1} & 57.1 & \textbf{59.7} & 67.6 & 50.9 & \textbf{71.3} \\ 
   \midrule 
SD v2~\cite{rombach2022high} &  80.8 & 68.6 & 50.0 & 63.6 & 72.3 & 63.6 & \textbf{69.3} & 62.9 & \textbf{61.9} & 55.7 & 66.8 & 60.9 & 70.3\\ 
+ \methodname{}  & \textbf{88.0} & \textbf{80.8} & \textbf{65.4} & \textbf{71.0} & \textbf{78.7} & \textbf{71.5} & 66.3 & \textbf{78.4} & 61.0 & \textbf{69.2} & \textbf{81.4} & \textbf{67.4} & \textbf{77.7}  \\
    \midrule
    SDXL~\cite{podell2023sdxl} & 84.8 & 74.7 & 58.0 & 69.4 & 71.1 & 60.7 & 59.8 & 71.7 & \textbf{61.5} & 63.9 & 71.7 & 60.0 & 73.3 \\ 
    + \methodname{} & \textbf{90.4} & \textbf{81.3} & \textbf{64.0} & \textbf{77.4} & \textbf{83.3} & \textbf{68.2} & \textbf{73.0} & \textbf{79.4} & 60.2 & \textbf{77.3} & \textbf{75.4} & \textbf{76.4} & \textbf{80.2} \\
    \bottomrule
  \end{tabular}
  }
\end{table}

\section{Skill-specific VQA Accuracy on TIFA and DSG} \label{sec:skill-specific-acc}
In this section, we show the detailed VQA accuracy for each skill category on TIFA and DSG benchmarks.
Since DreamSync~\cite{sun2023dreamsync} does not provide the skill-specific scores, we report the skill-specific scores of SD models on TIFA and DSG and based on our experiments in \Cref{table:tifa_skill} and \Cref{table:dsg_skill};
we observe there are less than 1\% score differences of SD/SDXL models in TIFA average accuracy between the results in Sun~\etal{} and ours.

As shown in \Cref{table:tifa_skill} and \Cref{table:dsg_skill}, on both TIFA and DSG benchmarks, \methodname{} improves the generation faithfulness in most of the categories.
Comparing SDXL and SDXL+\methodname{},
the SDXL finetuned with \methodname{} approach shows large improvement especially in entity (\ie{} 3.1\% on animal/human on TIFA compared with SDXL, 5.6\% in whole on DSG, 12.2\% in part on DSG),
as well as 
spatial relationship (\ie{} 1.9\% on TIFA, and 8.0\% on DSG),
and counting skills (\ie{} 1.4\% on TIFA, and 13.2\% on DSG).
Besides, we also observe that \methodname{} significantly improves the text rendering for SD v2 and SDXL (\ie{} 16.4\% compared with SDXL, and 6.5\% compared with SD v2 on DSG), but not SD v1.4 (\ie{} 1.8\% decrease compared with SD v1.4 on DSG).

\begin{figure}[t]
    \centering
    \includegraphics[width=.9\textwidth]{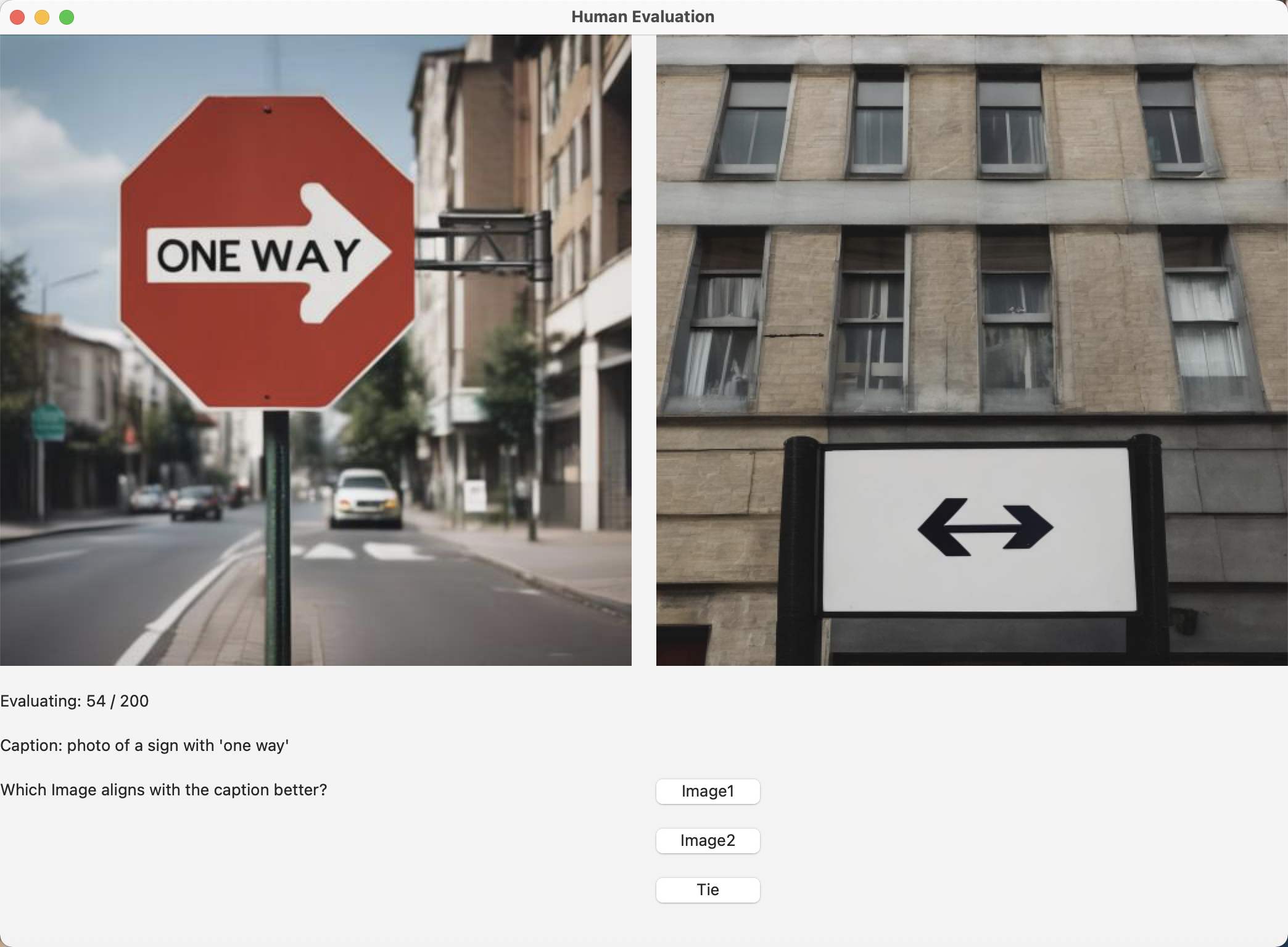}
    \caption{Example user interface for human evaluation on DSG prompts. 
    }
   \label{fig:evalui}
\end{figure}

\begin{wraptable}[12]{r}{0.4\textwidth}
\vspace{-20pt}
\caption{Human Evaluation on 200 sampled text prompts from DSG.
We show the detailed win/lose/tie counts on all samples and samples from each dataset.}
\label{table:human_eval_appendix}
\resizebox{0.4\columnwidth}{!}{
\begin{tabular}{lccc} \\
\toprule  
\textbf{Eval Dataset} &  \textbf{Win} & \textbf{Lose} & \textbf{Tie}  \\
\midrule
All & 241 & 114 & 245 \\ 
Localized Narratives & 19 & 5 & 12 \\ 
DiffusionDB & 6 & 5 & 34 \\
CountBench & 26 & 11 & 17 \\ 
COCO & 35 & 2 & 47 \\ 
Whoops & 21 & 10 & 26\\ 
\bottomrule
\end{tabular}
}
\end{wraptable}

\section{Human Evaluation Details}  \label{sec:detail_human_eval}
We conduct the human evaluation (described in the main paper \cref{sec:human_eval}) on 200 randomly sampled prompts from DSG, with three external annotators.
We show the annotation interface in \cref{fig:evalui}.
The image order between the two models is randomized to avoid the leakage of information about which image is generated with which model.

In \Cref{table:human_eval_appendix},
we show the detailed annotator votes for win, lose, and tie for
SDXL and SDXL+\methodname{}.
SDXL+\methodname{} has significantly higher win votes compared with SDXL on all the 200 sampled text prompts (\ie{}, 241 win \vs{} 114 lose), demonstrating the effectiveness of \methodname{}.

\begin{figure}[h]
    \centering
    \includegraphics[width=.8\textwidth]{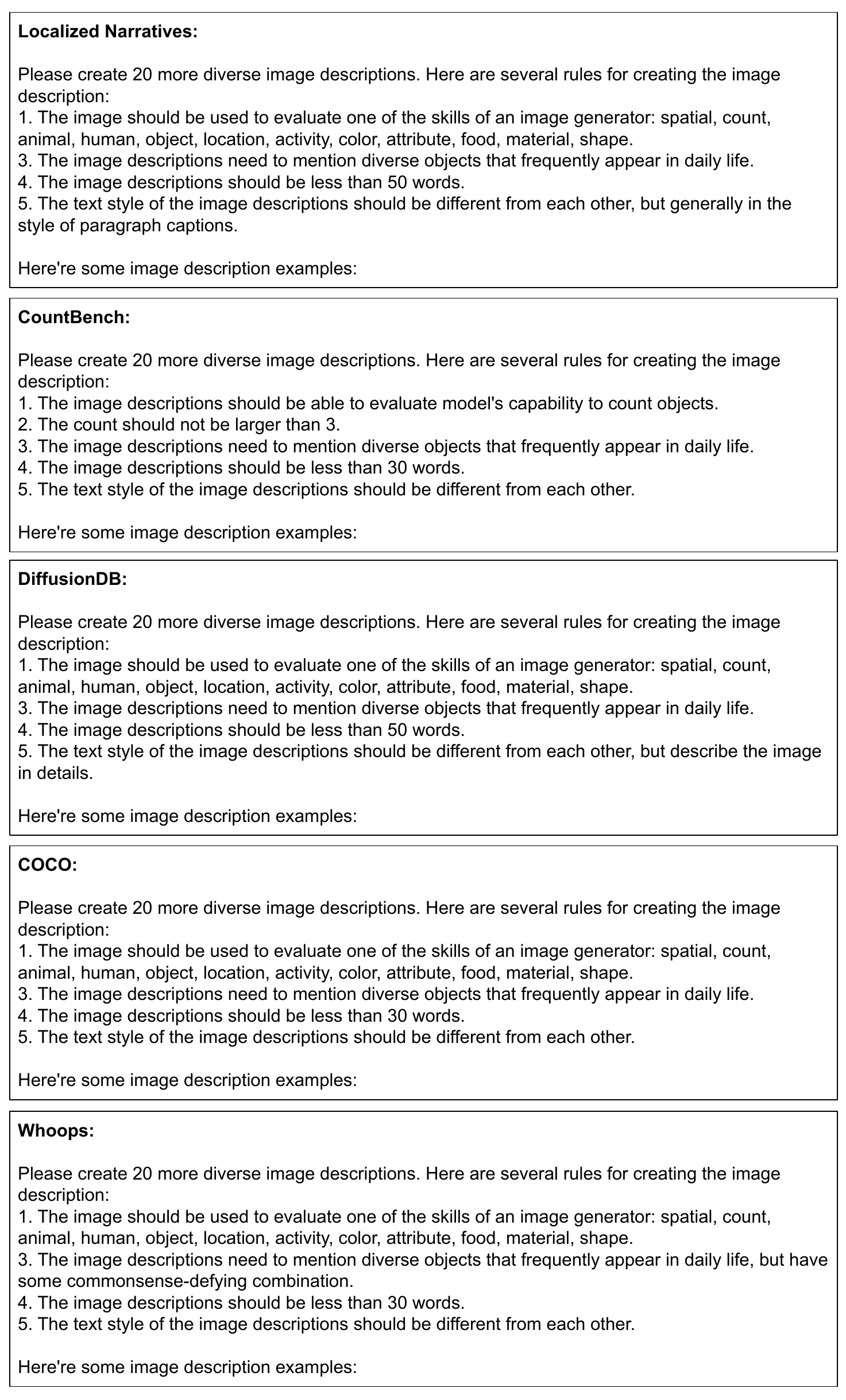}
    \caption{Prompts used to query GPT-3.5 auto-generate new prompts targeting different skills. 
    }
   \label{fig:gpt_prompt}
\end{figure}

\section{Implementation Details of DPO and MoE-LoRA} \label{sec:dpo_moelora_imp}

In this section, we provide the implementation details of two training approaches we experiment with
(described in \cref{sec:training_ablation} in the main paper). 

\noindent\textbf{Direct Preference Optimization (DPO).}
We fine-tune LoRA models with DPO proposed in~\cite{wallace2023diffusion}. Specifically, we sample two images with T2I models and calculate the image-text alignment with CLIP score~\cite{radford2021learning}. We use the image with a higher CLIP score as the positive example and the image with a lower CLIP score as the negative example. We fine-tune the LoRA models to learn to generate images closer to the positive image distribution and push away from generating images similar to the negative image distribution. Similarly, we fine-tune with DPO on five datasets with different text styles and skills and merge LoRA expert models during inference time by averaging the LoRA weights. In DPO training, we use a constant learning rate 3e-4 and fine-tune LoRA for 5K steps. We evaluate the model on DSG every 1K steps and pick the best checkpoint.

\noindent\textbf{Mixture of Lora Experts (MoE-LoRA).}
MoE-LoRA~\cite{wu2023mole} utilizes a gating function (router) to decide which experts to use during training and inference. The gating function predicts weights for each expert based on layer inputs and picks the top $K$ experts to use at each layer. Specifically, the gating function we use is a simple linear mapping function, where $\{w_i\}_{i=1}^{K} = W_gx$. $x$ is the input to each layer, $W_g$ is the learnable gating weights, and $\{w_i\}_{i=1}^{K}$ are the predicted weights. The outputs of each expert are added together with the normalized weights from the gating function. In MoE-LoRA, we initialize five LoRA experts fine-tuned on different datasets containing different text styles and skills. We freeze the learned LoRA weights and only fine-tune the gating function on the collected five datasets. We activate all five experts during training and inference (\ie{}, $K=5$). In MoE-LoRA training, we use a constant learning rate 1e-5 and fine-tune LoRA for 5K steps. We evaluate the model on DSG every 1K steps and pick the model with highest text faithfulness score.

\section{Skill-Specific Prompt Generation Details} \label{sec:prompt_detail}

We show the prompts we use to query GPT3.5 to generate 1K prompts for each skill.
As shown in \cref{fig:gpt_prompt}, we use different prompts to generate \methodname{} data.
For example, we specify ``paragraph captions'' to generate text prompts that can be used to teach model to follow long text prompts, and specifying ``evaluate model's capability to count objects'' to collect a set of prompts for improving model's counting capability.
Besides, in all the prompt generation, we emphasize that the image should ``mention diverse objects'' to maximize the semantic diversity in generated prompts.

\section{Limitations}
\label{sec:limitation}
\methodname{} relies on a strong image generator
and an instruction-following LLM.
Note that \methodname{} is model-agnostic and can be implemented with publicly accessible models (GPT-3.5 and Stable Diffusion models).
Also, since our fine-tuning works well with a small number of image-text pairs
(\ie{}, for each skill, we only generate 1K text prompts and generating one image per each text prompt),
the cost of LLM inference (\ie{}, \$27.78 for querying GPT-3.5 for generating prompts in all the experiments including ablation studies) and image generation (8s per image for image generation with SDXL on a single NVIDIA L40 GPU with 48GB memory) is minimal.
Second,
although \methodname{} helps boost T2I models' performance significantly in both text faithfulness and alignment to human preference, 
fine-tuning with \methodname{} 
does not 
guarantee the resulting model to follow the text prompts in every detail.
To use T2I models trained with \methodname{}, researchers should first carefully study their capabilities in relation to the specific context they are being applied within.